\documentclass[11pt, a4paper]{article}

\usepackage[utf8]{inputenc}
\usepackage[T1]{fontenc}
\usepackage{lmodern}
\usepackage[margin=1in]{geometry}
\usepackage{amsmath, amsfonts, amssymb}
\usepackage{booktabs}
\usepackage{graphicx}
\usepackage{subcaption}
\usepackage{cite}
\usepackage{hyperref}
\usepackage{xcolor}

\hypersetup{
    colorlinks=true,
    linkcolor=blue!70!black,
    citecolor=green!50!black,
    urlcolor=blue!80!black
}

\title{\textbf{\Large A Neurosymbolic Multimodal Framework for 30-Day Diabetic Readmission Prediction: Integrating Structured EHR Data, Dense Clinical Embeddings, and Knowledge-Grounded Symbolic Extraction}}

\author{
    \textbf{Una Suman Kumar Patro}\textsuperscript{1,*}, \textbf{Sibangi Subhadarsani}\textsuperscript{2}, and \textbf{Narendra Badam}\textsuperscript{3} \\[0.6em]
    \textsuperscript{1,2,3}\textit{Independent Researcher, Celina, TX 75009, USA} \\[0.4em]
    \small Email: \texttt{samsumankumarpatro@gmail.com}, \texttt{sibangisubhadarsani@gmail.com}, \\
    \small \texttt{narendrabadam4@gmail.com} \\[0.2em]
    \small \textsuperscript{*}Corresponding author: Una Suman Kumar Patro (\texttt{samsumankumarpatro@gmail.com}) \\[0.3em]
    \small ORCID: Una Suman Kumar Patro \href{https://orcid.org/0009-0006-4639-7359}{0009-0006-4639-7359}; \\
    \small Sibangi Subhadarsani \href{https://orcid.org/0009-0003-1794-5684}{0009-0003-1794-5684}; Narendra Badam \href{https://orcid.org/0009-0004-9676-8053}{0009-0004-9676-8053}
}
\date{}

\begin{document}

\maketitle

\begin{abstract}
\noindent Unplanned 30-day readmissions among patients with diabetes mellitus impose a substantial clinical and economic burden, yet most predictive models use only structured electronic health record (EHR) data and ignore free-text clinical notes, which often contain code-mixed Hindi--English (Hinglish) phrasing and negation that general-purpose language models parse unreliably. We propose a neurosymbolic multimodal framework evaluated on the public Diabetes 130-US Hospitals dataset ($101\,766$ encounters; $71\,518$ patients) with a patient-level grouped 80/20 split. Three streams are fused: engineered structured features; 32 components of \texttt{Bio\_ClinicalBERT} note embeddings obtained by truncated singular value decomposition; and symbolic entities extracted by an open-weight clinical language model within a three-stage hybrid retrieval-augmented generation pipeline (dense retrieval, sparse lexical retrieval and cross-encoder re-ranking) with reflective negation auditing. A cost-sensitive \texttt{XGBoost} classifier with a decision threshold calibrated by Youden's $J$ statistic produces the prediction. On the held-out test cohort ($n = 20\,153$), the structured baseline achieved an area under the receiver operating characteristic curve (AUROC) of $0.6477$, sensitivity of $0.5839$ and Brier score of $0.1080$. The full multimodal model achieved the highest sensitivity ($0.6667$; $+8.28$ percentage points), the best $F_2$-score ($0.3824$) and the lowest Brier score ($0.0977$), at a slightly lower AUROC ($0.6375$) and specificity ($0.5161$). Shapley additive explanations showed that prior utilization dominates risk, with compressed narrative components among the top predictors. On a 500-note code-switched benchmark, grounded retrieval raised Hinglish symptom recall from $4.8\%$ to $40.2\%$ and precision from $3.1\%$ to $45.2\%$, while reducing entity hallucinations from $1.74$ to $0.00$ per note.

\vspace{0.8em}
\end{abstract}

\section{Introduction}
\label{sec:introduction}

Diabetes mellitus currently affects more than 537 million adults worldwide; projections indicate a rise to 783 million by 2045 \cite{strack2014impact}. Beyond chronic microvascular and macrovascular sequelae, acute metabolic decompensations drive a high volume of emergency hospitalizations and unplanned 30-day readmissions. In the United States the Centers for Medicare \& Medicaid Services Hospital Readmissions Reduction Program financially penalizes facilities with excess readmission rates, thereby creating a pressing need for prospective, well-calibrated clinical decision-support systems that can guide post-discharge transitional care \cite{hansen2011preventable}.

Two fundamental obstacles continue to limit the performance of contemporary readmission models. First, the majority of published risk tools rely almost exclusively on structured electronic health record (EHR) matrices---demographic attributes, laboratory values, diagnosis codes and medication counts \cite{strack2014impact, rajkomar2018scalable}. While these fields capture discrete physiological measurements, they systematically omit contextual information recorded in free-text clinical notes, including symptom trajectory, functional status and social determinants of health \cite{garcia2020multimodal}. Second, clinical documentation in diverse healthcare environments frequently incorporates colloquial expressions, non-standard abbreviations and code-mixed phrasing (for example Hinglish) that standard English medical dictionaries fail to recognize \cite{mishra2023code, kearney2024mitigating}. Concurrently, large language models (LLMs) exhibit strong linguistic reasoning yet, when deployed as end-to-end prognosticators, introduce risks of factual hallucination, poorly calibrated probability estimates and systematic failures in medical negation resolution \cite{chapman2001simple, diaz2025optimized}.

\begin{figure}[htbp]
\centering
\includegraphics[width=0.95\textwidth]{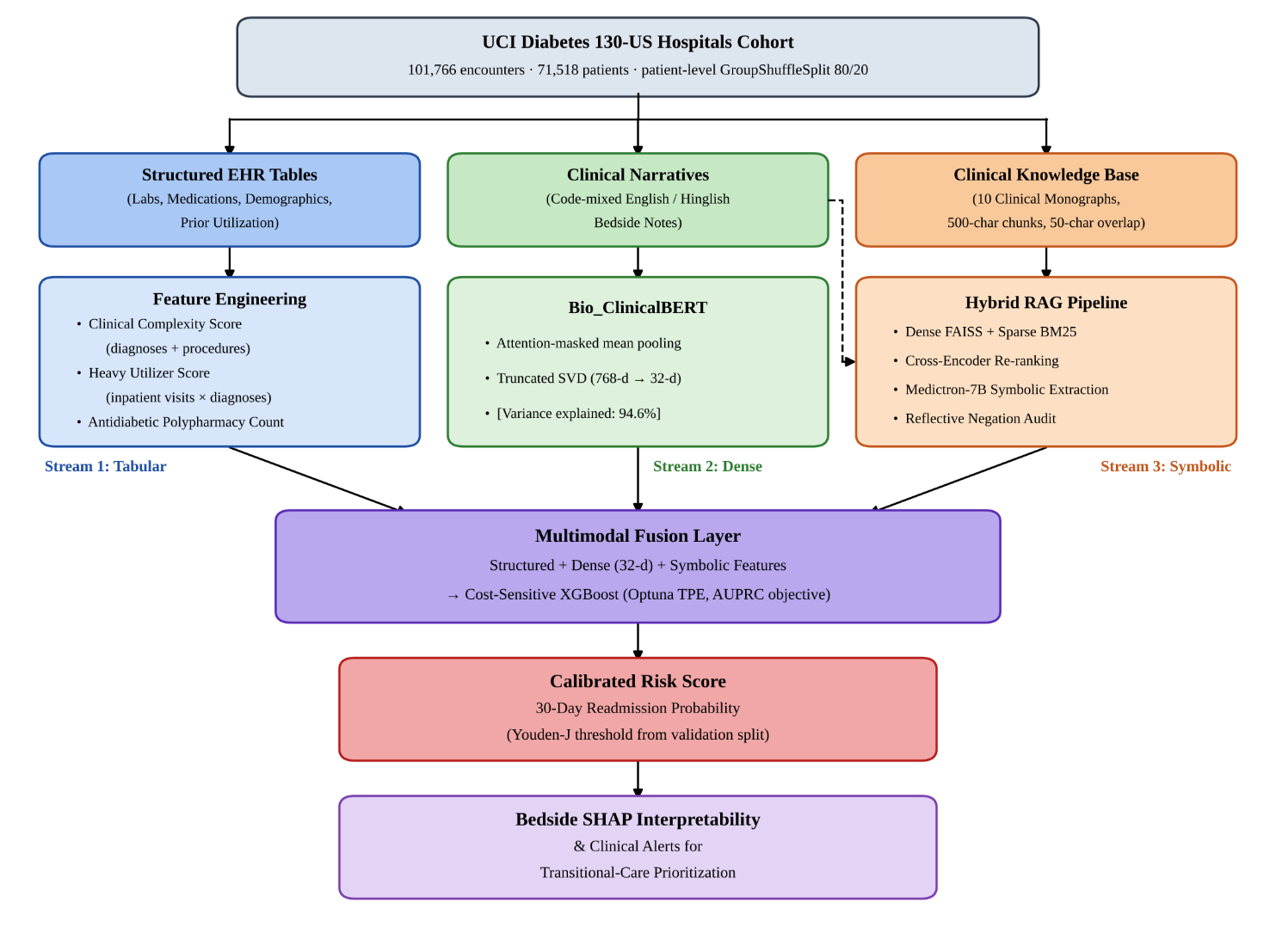}
\caption{End-to-end neurosymbolic multimodal architecture. Structured EHR features, SVD-compressed \texttt{Bio\_ClinicalBERT} embeddings ($768\text{-d} \rightarrow 32\text{-d}$) and symbolic entities extracted by the hybrid RAG pipeline (the clinical note serves as the retrieval query) are fused in a cost-sensitive \texttt{XGBoost} classifier whose calibrated 30-day readmission probability is delivered with bedside SHAP explanations.}
\label{fig:architecture}
\end{figure}

To address these limitations we present a neurosymbolic multimodal architecture (Fig.~\ref{fig:architecture}) that integrates three complementary information streams: (i) engineered structured EHR features, (ii) dense contextual representations obtained from \texttt{Bio\_ClinicalBERT} and compressed by truncated singular-value decomposition (SVD), and (iii) discrete symbolic clinical entities extracted by an open-weight biomedical language model (\texttt{Medictron-7B}) operating inside a hybrid retrieval-augmented generation (RAG) framework \cite{alsentzer2019publicly, reddy2024medictron, labrak2024biomistral, lundberg2017unified}.

\subsection{Key contributions}
\begin{enumerate}
    \item A unified pipeline that concatenates structured tabular variables, 32-dimensional SVD-compressed \texttt{Bio\_ClinicalBERT} embeddings and discrete symbolic indicators into a cost-sensitive \texttt{XGBoost} classifier.
    \item A three-stage retrieval engine---Facebook AI Similarity Search (FAISS) dense search, Okapi best-matching 25 (BM25) sparse retrieval and cross-encoder re-ranking---grounded on a curated clinical knowledge base of ten monographs, coupled with a reflective extraction loop that standardizes code-mixed terms and separates affirmed from negated symptoms.
    \item Systematic evaluation of continuous dense embeddings against discrete symbolic features inside tree-based classifiers under severe class imbalance, quantifying effects on sensitivity, discrimination and probability calibration.
    \item SHapley Additive exPlanations (SHAP) analysis demonstrating how narrative-derived features resolve risk factors that remain invisible to structured tabular predictors alone.
    \item Patient-level grouped partitioning that eliminates identity leakage and adherence to the Transparent Reporting of a multivariable prediction model for Individual Prognosis Or Diagnosis (TRIPOD) and Minimum Information about CLinical Artificial Intelligence Modeling (MI-CLAIM) reporting standards \cite{collins2015transparent, norgeot2020minimum}.
\end{enumerate}

\section{Related Work}
\label{sec:related_work}

Predictive modeling on electronic health records has matured rapidly around public benchmarks such as the University of California, Irvine (UCI) Diabetes 130-US Hospitals dataset \cite{strack2014impact} and MIMIC-III \cite{johnson2016mimic}. Early studies employed logistic regression and classical tree ensembles \cite{rajkomar2018scalable}. Gradient-boosted decision trees, particularly \texttt{XGBoost} \cite{chen2016xgboost}, have demonstrated strong empirical performance on tabular clinical data because of their capacity to model non-linear interactions and to handle missing values natively. Nevertheless, models restricted to structured tables cannot exploit the rich narrative context present in discharge summaries and progress notes. More recent deep-learning approaches that process longitudinal sequences of diagnosis and procedure codes have improved discrimination on some cohorts, yet they still discard the free-text layer of the medical record.

Domain-specific pretrained transformers such as \texttt{BioBERT} \cite{lee2020biobert} and \texttt{Bio\_ClinicalBERT} \cite{alsentzer2019publicly, huang2019clinicalbert} produce contextual token representations well suited to clinical named-entity recognition and relation extraction. Direct concatenation of the native 768-dimensional embeddings with tabular features, however, has been shown to diffuse decision boundaries inside tree ensembles and to inflate false-positive rates under class imbalance \cite{diaz2025optimized}. Dimensionality reduction prior to fusion is therefore essential. Truncated SVD offers a computationally efficient linear projection that preserves the dominant semantic directions while yielding a compact representation compatible with axis-aligned splits.

Large language models have demonstrated utility on clinical reasoning tasks \cite{singhal2023large}, yet direct deployment is constrained by hallucination risk and computational cost. Retrieval-augmented generation \cite{lewis2020retrieval} mitigates factual errors by conditioning generation on verified reference passages. Open-weight biomedical models, including \texttt{BioMistral-7B} \cite{labrak2024biomistral} and \texttt{Medictron-7B} \cite{reddy2024medictron}, enable local, privacy-preserving structured extraction without transmission of protected health information to external application programming interfaces (APIs). Hybrid dense-plus-sparse retrieval further improves recall of rare clinical abbreviations and code-switched expressions.

Multimodal learning seeks to synthesize heterogeneous clinical data modalities---tabular measurements, physiologic waveforms and unstructured notes \cite{garcia2020multimodal}. Neurosymbolic approaches integrate continuous representation learning with discrete logical symbols, thereby improving interpretability and providing natural guard-rails \cite{nelson2022explainable}. In the present work language models function exclusively as symbolic parsers that convert free-text into standardized categorical and Boolean variables for subsequent gradient boosting. This design choice deliberately avoids treating the LLM as a black-box risk predictor and instead exploits its linguistic competence while retaining the calibration and ranking strengths of tree ensembles.

\section{Materials and Methods}
\label{sec:materials_and_methods}

\subsection{Dataset provenance and cohort definition}
We utilized the Diabetes 130-US Hospitals dataset (1999--2008) released through the UCI Machine Learning Repository \cite{strack2014impact}. This publicly available multicenter cohort was originally assembled from the Health Facts database and has become a de-facto benchmark for diabetic readmission modeling because of its size, geographic diversity and rich mixture of administrative, laboratory and medication fields. The cohort comprises
\begin{equation}
N = 101\,766
\label{eq:N_encounters}
\end{equation}
inpatient encounters. Each record corresponds to a discrete hospital stay and is linked to a unique patient identifier, thereby enabling longitudinal tracking of repeated admissions. The encounters correspond to
\begin{equation}
P = 71\,518
\label{eq:P_patients}
\end{equation}
unique patients treated across 130 medical centers spanning both academic and community hospitals in the United States. Inclusion required an established diagnosis of diabetes mellitus (International Classification of Diseases, Ninth Revision [ICD-9] codes 250.xx), a hospital length of stay between 1 and 14 days (thereby excluding ultra-short observational stays and prolonged complex hospitalizations), laboratory testing during the encounter, and administration of at least one antidiabetic agent. These criteria ensure that every observation represents a clinically meaningful inpatient episode of care for a patient whose diabetes was actively managed during the hospitalization. The primary outcome is binary 30-day hospital readmission, formally defined as
\begin{equation}
y_i = 
\begin{cases} 
1, & \text{if patient } i \text{ is readmitted within 30 days}, \\
0, & \text{otherwise}.
\end{cases}
\label{eq:outcome}
\end{equation}
Baseline characteristics of the cohort and of both partitions are summarized in Table~\ref{tab:cohort_characteristics}. The positive class prevalence is approximately $11.2\%$, confirming severe class imbalance. In such settings the na\"ive accuracy maximizer is strongly biased toward the majority (non-readmitted) class; consequently, discrimination metrics that are insensitive to prevalence (area under the receiver operating characteristic curve, AUROC) must be complemented by precision-recall analysis, cost-sensitive learning and careful threshold calibration. The imbalance also motivates the use of stratified sampling and class-weighted loss functions in all subsequent modeling stages.

\begin{table}[htbp]
\centering
\caption{Demographic and baseline clinical characteristics of the study cohort and of the patient-level training and test partitions.}
\label{tab:cohort_characteristics}
\begin{tabular}{@{}lccc@{}}
\toprule
\textbf{Characteristic} & \textbf{Full cohort} & \textbf{Training} & \textbf{Test} \\
\midrule
Encounters, $n$ & $101\,766$ & $81\,613$ & $20\,153$ \\
Unique patients, $n$ & $71\,518$ & $57\,214$ & $14\,304$ \\
Age $< 30$ y, $n$ (\%) & $2\,509$ (2.47) & $2\,018$ (2.47) & $491$ (2.44) \\
Age 30--49 y, $n$ (\%) & $13\,460$ (13.23) & $10\,790$ (13.22) & $2\,670$ (13.25) \\
Age 50--69 y, $n$ (\%) & $39\,739$ (39.05) & $31\,799$ (38.96) & $7\,940$ (39.40) \\
Age 70--89 y, $n$ (\%) & $43\,265$ (42.51) & $34\,805$ (42.65) & $8\,460$ (41.98) \\
Age $\ge 90$ y, $n$ (\%) & $2\,793$ (2.74) & $2\,201$ (2.70) & $592$ (2.94) \\
Female, $n$ (\%) & $54\,708$ (53.76) & $43\,834$ (53.71) & $10\,874$ (53.96) \\
Length of stay, days, mean $\pm$ SD & $4.40 \pm 2.98$ & $4.40 \pm 2.99$ & $4.39 \pm 2.97$ \\
Laboratory procedures, mean $\pm$ SD & $43.10 \pm 19.67$ & $43.08 \pm 19.68$ & $43.15 \pm 19.66$ \\
Medications, mean $\pm$ SD & $16.02 \pm 8.13$ & $16.02 \pm 8.13$ & $16.03 \pm 8.14$ \\
Diagnoses, mean $\pm$ SD & $7.42 \pm 1.93$ & $7.43 \pm 1.93$ & $7.40 \pm 1.95$ \\
Prior inpatient visits, mean $\pm$ SD & $0.64 \pm 1.26$ & $0.64 \pm 1.27$ & $0.63 \pm 1.25$ \\
30-day readmission, $n$ (\%) & $11\,357$ (11.16) & $9\,206$ (11.28) & $2\,151$ (10.67) \\
\bottomrule
\multicolumn{4}{@{}l}{\footnotesize SD: standard deviation.}
\end{tabular}
\end{table}

\subsection{Patient-aware data partitioning}
A fundamental threat to the validity of any patient-level predictive model is identity leakage: when multiple encounters belonging to the same individual appear in both the training and test partitions, the model can memorize patient-specific patterns rather than learning generalizable risk factors. To preclude this form of data leakage, a grouped shuffle split (80/20, fixed random seed) using the patient identifier as the grouping key was applied. Formally, the partition satisfies
\begin{equation}
|\mathcal{D}_{\text{train}}| = 81\,613, \quad |\mathcal{D}_{\text{test}}| = 20\,153, \quad \operatorname{patient}(\mathcal{D}_{\text{train}}) \cap \operatorname{patient}(\mathcal{D}_{\text{test}}) = \emptyset
\label{eq:partition}
\end{equation}
All hyper-parameter tuning was performed exclusively inside the training partition using stratified three-fold cross-validation, and the decision threshold was calibrated on a stratified $20\%$ validation subset held out from the training partition; the independent test partition was examined only once for final reporting. This strict separation guarantees that reported performance estimates are unbiased with respect to patient identity and that no information from the test set influenced model development. The resulting training partition contained $81\,613$ encounters and the independent test partition contained $20\,153$ encounters (total $101\,766$). Because the split is grouped by patient rather than stratified by outcome, the readmission prevalence differs marginally between partitions (Table~\ref{tab:cohort_characteristics}).

\subsection{Structured feature engineering}
From the tabular EHR fields a compact set of clinically motivated composite features was derived. Raw counts of diagnoses, procedures and prior encounters are informative individually, but they do not capture the joint burden of comorbidity, procedural intensity and repeated acute-care use. The engineered features described below address these limitations while remaining fully transparent and reproducible; each is defined exactly as implemented in the released preprocessing code.

\paragraph{Clinical Complexity Score (CCS).}
The CCS is an additive measure of the diagnostic and procedural burden of an encounter, obtained by summing the number of recorded diagnoses and the number of procedures performed during the stay. High values indicate that many coexisting problems and interventions were documented within a single admission. Formally,
\begin{equation}
\text{CCS}_i = n_i^{\text{diag}} + n_i^{\text{proc}}
\label{eq:ccs}
\end{equation}
The score is deliberately unnormalized: length of stay enters the model as a separate predictor, so the tree ensemble can learn any interaction between complexity and duration directly rather than through a fixed ratio.

\paragraph{Heavy Utilizer Score (HUS).}
Prior healthcare utilization is among the strongest predictors of future readmission. The HUS is an interaction term that multiplies the number of inpatient admissions in the preceding year by the number of diagnoses recorded for the index encounter, thereby emphasizing patients who combine repeated hospitalization with high comorbidity burden. The score is zero for every patient without a prior inpatient admission. It is defined as
\begin{equation}
\text{HUS}_i = n_i^{\text{inpatient}} \times n_i^{\text{diag}}
\label{eq:hus}
\end{equation}
Because the score is a product rather than a sum, it grows rapidly for multimorbid frequent admitters (for example, seven prior admissions and five diagnoses yield $\text{HUS} = 35$; see Fig.~\ref{fig:waterfall_high_risk}). The HUS enters the model as a single numeric feature alongside its constituent counts.

\paragraph{Antidiabetic Polypharmacy Count (APC).}
Contemporary diabetes management frequently involves combination regimens. While polypharmacy can improve glycemic control, it also increases the risk of adverse drug events, therapeutic duplication and medication non-adherence---all of which may precipitate readmission. In the UCI extract each of 23 antidiabetic agents (biguanides, sulfonylureas, meglitinides, thiazolidinediones, alpha-glucosidase inhibitors, insulin and fixed-dose combinations) is recorded as a categorical status: \texttt{No}, \texttt{Steady}, \texttt{Up} or \texttt{Down}. The APC counts the agents whose status indicates active use during the encounter (\texttt{Steady}, \texttt{Up} or \texttt{Down}); a status of \texttt{No} contributes zero. The feature is labelled ``Total Meds'' in Figs.~\ref{fig:shap_bar}--\ref{fig:waterfall_low_risk}. Formally,
\begin{equation}
\text{APC}_i = \sum_{m=1}^{23} \mathbb{I}\left(s_{m,i} \in \{\text{Steady}, \text{Up}, \text{Down}\}\right)
\label{eq:apc}
\end{equation}
Age, which the UCI dataset reports in ten-year brackets, was mapped from the lower bound of each bracket into six strata: Pediatric ($<18$~y), Young Adult (18--34~y), Adult (35--49~y), Middle-Aged (50--64~y), Senior (65--79~y) and Elderly ($\ge 80$~y). Discretization permits the tree ensemble to learn non-monotonic age effects. The 23 individual medication-status columns (e.g., metformin, insulin) were retained as categorical features and the \texttt{diabetesMed} flag was binarized. Within the training pipeline, numeric features were median-imputed and categorical features were mode-imputed and ordinal-encoded; all imputation statistics were estimated on the training partition only.

\subsection{Grounded clinical note generation}
To evaluate cross-lingual and code-mixed extraction performance without compromising patient privacy or violating institutional data-governance policies, we implemented a controlled, attribute-conditioned clinical note generator that produces realistic bedside narratives deterministically mapped to the structured attributes of each encounter.

Real free-text notes from the original Health Facts extract are not publicly redistributable. Consequently, a synthetic generation strategy was adopted that preserves the statistical relationships between diagnosis categories, laboratory findings, medication regimens and documented symptoms while introducing controlled linguistic challenges. For every encounter a short progress-note style narrative (approximately 80--150 words) was produced. Generation was conditioned on the primary diagnosis category (\texttt{diag\_1\_category}), the derived glucose-status label, the presence of key comorbidities, and a randomly sampled subset of active antidiabetic agents. This conditioning guarantees that the semantic content of the note remains clinically coherent with the structured record; for example, an encounter whose primary diagnosis falls in the circulatory category is far more likely to contain mentions of chest pain or dyspnea than an encounter dominated by a soft-tissue infection.

A central design goal was to stress-test the downstream symbolic extraction pipeline under realistic multilingual conditions. Therefore the generator deliberately injects code-switched Hindi--English (Hinglish) expressions that are commonly observed in South-Asian clinical documentation and in the speech of bilingual patients and providers. Representative lexical substitutions include:
\begin{itemize}
    \item \textit{``seene me dard''} / \textit{``seene mein dard''} $\rightarrow$ chest pain
    \item \textit{``saans lene me dikkat''} / \textit{``saans phoolna''} $\rightarrow$ dyspnea / shortness of breath
    \item \textit{``thakan aur kamzori''} $\rightarrow$ fatigue and weakness
    \item \textit{``tez bukhar''} $\rightarrow$ high-grade fever
    \item \textit{``pairon mein sunnapan''} $\rightarrow$ numbness in the feet
    \item \textit{``bahut pyaas''} $\rightarrow$ polydipsia
\end{itemize}
In addition to code-mixing, the generator systematically inserts both affirmed and explicitly negated findings. Negation is realized through a mixture of English markers (\textit{``denied''}, \textit{``no''}, \textit{``absent''}, \textit{``patient reports no \dots''}) and Hindi/Hinglish markers (\textit{``nahi''}, \textit{``nahin''}, \textit{``bilkul nahi''}, \textit{``mana kiya''}). Typical constructions include:
\begin{itemize}
    \item \textit{``Severe radiating chest pain bilkul nahi hai.''}
    \item \textit{``Foot numbness denied; monofilament sensation intact.''}
    \item \textit{``Patient complains of saans lene me dikkat but denies orthopnea.''}
\end{itemize}
The co-occurrence of affirmed and negated symptoms within the same note creates a non-trivial disambiguation task for any extraction model and directly motivates the reflective negation-auditing stage described in Section~\ref{sec:symbolic_extraction}.

Because the mapping from structured attributes to narrative content is deterministic given a fixed random seed, the generated notes are fully reproducible. At the same time, the controlled injection of code-switching and negation yields a challenging yet privacy-preserving benchmark on which the contribution of the hybrid RAG pipeline and the multilingual negation detector can be quantified (see Section~\ref{sec:hinglish_benchmark}). While synthetic notes cannot capture every pragmatic and stylistic nuance of authentic clinical documentation, they provide a transparent, ethically unencumbered test bed for evaluating the robustness of neurosymbolic extraction under the precise linguistic phenomena the architecture is designed to address.

\subsection{Three-stage hybrid RAG architecture}
A knowledge repository comprising ten concise clinical monographs was constructed to serve as the external memory for retrieval-augmented extraction. The monographs cover diabetes pathophysiology, glycated hemoglobin (HbA1c) interpretation, diabetic ketoacidosis, clinical indicators of acute glycemic volatility, microvascular complications, oral antidiabetic drugs, medication interactions, readmission risk factors, hospital quality metrics, and a bilingual (English--Hinglish) clinical glossary of symptom descriptors and colloquial expressions. Each monograph was written in a concise, textbook-style register (approximately 300--700 characters) so that retrieved passages remain factually dense and free of narrative digression.

\begin{figure}[htbp]
\centering
\includegraphics[width=0.95\textwidth]{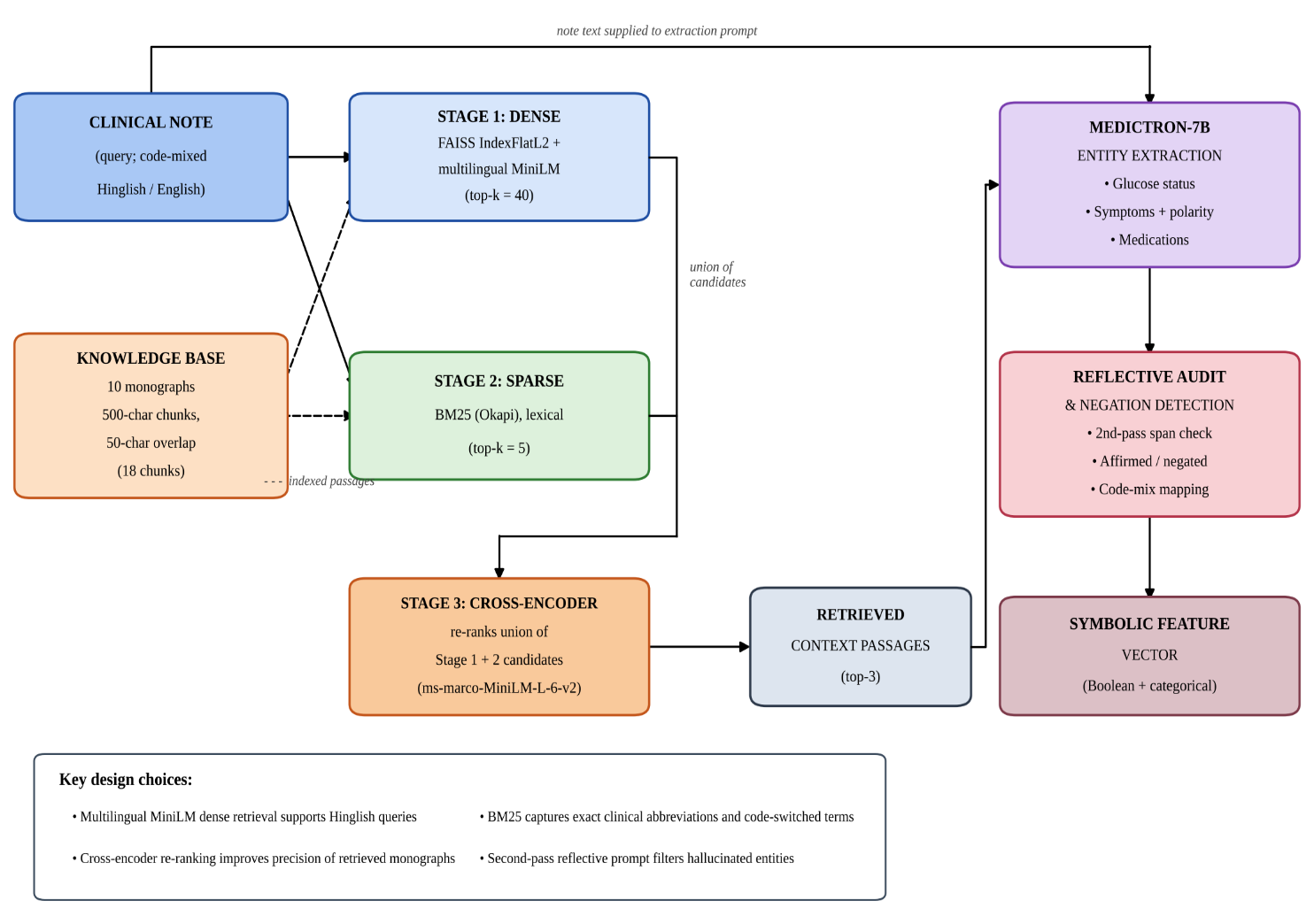}
\caption{Three-stage hybrid retrieval-augmented generation pipeline: dense FAISS ($L_2$) retrieval and BM25 sparse matching produce a candidate union that is re-ranked by a cross-encoder; the top-three passages condition \texttt{Medictron-7B}, whose output passes a reflective audit and multilingual negation detection before emission as symbolic features.}
\label{fig:hybrid_rag}
\end{figure}

Documents were segmented into 500-character chunks with a 50-character overlap between consecutive chunks (stride 450 characters), yielding 18 indexed passages. The relatively short chunk length keeps individual retrieval units within the effective context window of the downstream cross-encoder while preserving local semantic coherence; the $10\%$ overlap mitigates the risk that a clinically critical sentence is split across chunk boundaries. The chunk collection was indexed by a three-stage hybrid retrieval cascade (Fig.~\ref{fig:hybrid_rag}) that deliberately combines the complementary strengths of dense semantic matching, sparse lexical matching, and pairwise re-ranking.

\paragraph{Dense retrieval.}
Queries (the clinical note, or a condensed symptom-oriented projection of the note) and all knowledge-base passages are encoded by the multilingual sentence transformer \texttt{paraphrase-multilingual-MiniLM-L12-v2}. This model was selected because it produces 384-dimensional embeddings that are effective for both English and code-mixed Hindi--English text, thereby supporting cross-lingual retrieval without requiring a separate translation step. The embeddings are stored in an exact FAISS \cite{johnson2019faiss} \texttt{IndexFlatL2} index, and candidate passages are ranked by ascending squared Euclidean ($L_2$) distance between the query embedding and each passage embedding:
\begin{equation}
d(q, p) = \|\mathbf{e}_q - \mathbf{e}_p\|_2^2
\label{eq:l2_distance}
\end{equation}
The top-$k = 40$ passages with the smallest $L_2$ distance are returned ($k = 4 \times 10$; with the current 18-passage index the entire knowledge base is ranked). Dense retrieval excels at capturing paraphrastic and semantically related content (e.g., matching ``shortness of breath'' to a passage discussing ``dyspnea on exertion''), yet it can under-emphasize rare technical abbreviations or exact code-switched tokens that appear infrequently in the pre-training corpus of the encoder.

\paragraph{Sparse retrieval.}
To compensate for the limitations of pure dense matching, a parallel BM25 index is maintained over the same chunk collection. BM25 \cite{robertson2009probabilistic} is a probabilistic lexical ranking function that rewards term frequency while applying a saturation effect and a length-normalization penalty:
\begin{equation}
\operatorname{BM25}(q, d) = \sum_{t \in q} \operatorname{IDF}(t) \cdot \frac{f(t, d) \cdot (k_1 + 1)}{f(t, d) + k_1 \cdot \left(1 - b + b \cdot \frac{|d|}{\operatorname{avgdl}}\right)}
\label{eq:bm25}
\end{equation}
where $f(t, d)$ is the frequency of term $t$ in document $d$, $|d|$ is the document length, $\operatorname{avgdl}$ is the average document length in the collection, and $k_1$ and $b$ are free parameters (set to the conventional defaults $1.5$ and $0.75$). Because BM25 operates directly on surface forms, it reliably surfaces passages that contain exact Hinglish symptom phrases or drug-name variants even when those strings are poorly represented in the dense embedding space. The top-5 BM25 hits are retained.

\paragraph{Cross-encoder re-ranking.}
The union of the dense and sparse candidate sets (duplicates removed) is then passed to a cross-encoder, \texttt{ms-marco-MiniLM-L-6-v2}. Unlike the bi-encoder used for dense retrieval, the cross-encoder concatenates the query and each candidate passage into a single sequence and produces a calibrated relevance score via a classification head. This joint encoding captures fine-grained interactions (negation scope, qualifier attachment, synonymy) that are invisible to independent embedding similarity. After re-ranking, only the three highest-scoring passages are retained; these three passages constitute the final context window that is prepended to the extraction prompt supplied to \texttt{Medictron-7B}.

The three-stage design therefore implements a classic ``recall-then-precision'' cascade: dense and sparse retrieval maximize coverage of potentially relevant knowledge, while the cross-encoder enforces high precision before the comparatively expensive language-model generation step is invoked. The benchmark in Section~\ref{sec:hinglish_benchmark} contrasts zero-shot extraction, unconstrained generative RAG and the grounded configuration, quantifying the contribution of retrieval and grounding to the recall and factual precision of the extracted symbolic features.

\subsection{Symbolic extraction and negation auditing}
\label{sec:symbolic_extraction}
Structured extraction is performed by \texttt{Medictron-7B}, an open-weight biomedical language model obtained by continued pre-training and instruction tuning of \texttt{BioMistral-7B} on a mixture of clinical guidelines, discharge summaries and structured extraction exemplars. Because the model weights remain fully local, protected health information never leaves the institutional boundary---an essential requirement for any production clinical natural language processing (NLP) pipeline.

Given a clinical note and the three passages retrieved by the hybrid RAG cascade, \texttt{Medictron-7B} is prompted to populate a fixed JSON schema that comprises:
\begin{itemize}
    \item demographic stratum (age group),
    \item ordinal glucose status (hyperglycemia / hypoglycemia / normoglycemia),
    \item a list of active antidiabetic medications,
    \item a list of symptom mentions, each accompanied by a binary polarity flag.
\end{itemize}
The schema-constrained generation format markedly reduces the incidence of free-form hallucination relative to unconstrained decoding. Nevertheless, even constrained models occasionally emit entities that cannot be aligned to any span in the source text. Consequently a second-pass reflective verification prompt is issued: the model is shown its own previous output together with the original note and is instructed to retain only those entities for which an explicit supporting span can be cited. Entities lacking a verifiable span are discarded.

A further, independent multilingual negation detector then operates on the verified symptom list. The detector maintains a lexicon of English and Hindi/Hinglish negation cues (\textit{``no''}, \textit{``not''}, \textit{``denied''}, \textit{``absent''}, \textit{``nahi''}, \textit{``nahin''}, \textit{``bilkul nahi''}, \dots) and applies a lightweight scope-resolution heuristic to determine whether each symptom falls inside a negated context. The result is a pair of complementary indicator variables for every symptom $j$:
\begin{equation}
s_j^+ = 1 \text{ if symptom } j \text{ is affirmed}, \qquad s_j^- = 1 \text{ if symptom } j \text{ is negated}.
\label{eq:negation_indicators}
\end{equation}
This explicit separation is clinically decisive. In a purely presence-based encoding the statement \textit{``patient denies chest pain''} would be indistinguishable from \textit{``patient reports chest pain,''} causing the downstream classifier to attribute elevated risk to a finding that the clinician has in fact ruled out. By contrast, the signed indicators allow the tree ensemble to learn that $s_j^-$ often exerts a protective (negative) SHAP contribution, thereby suppressing false-positive alerts.

\subsection{Dense representation stream}
In parallel with the symbolic pathway, the same unstructured narratives are encoded by \texttt{Bio\_ClinicalBERT}, a domain-adapted transformer pre-trained on PubMed abstracts and MIMIC-III clinical notes. \texttt{Bio\_ClinicalBERT} produces contextual token representations $\mathbf{h}_t \in \mathbb{R}^{768}$ that capture nuanced semantic relationships (e.g., the distinction between ``controlled'' and ``uncontrolled'' hyperglycemia) that discrete symbolic features necessarily abstract away.

Token-level vectors are aggregated into a single note-level embedding by attention-mask-weighted mean pooling:
\begin{equation}
\mathbf{e} = \frac{\sum_t \alpha_t \mathbf{h}_t}{\sum_t \alpha_t}, \qquad \alpha_t = \text{attention mask}_t.
\label{eq:mean_pooling}
\end{equation}
The attention mask ensures that padding tokens do not dilute the representation. The resulting 768-dimensional vectors, while rich, are problematic for tree-based classifiers: axis-aligned splits become statistically unstable in high-dimensional continuous spaces, especially under class imbalance, leading to fragmented decision boundaries and inflated variance. Truncated singular-value decomposition is therefore applied to project the embeddings onto a compact orthogonal subspace:
\begin{equation}
\mathbf{E}_{32} = \mathbf{U}_{32} \mathbf{\Sigma}_{32} \mathbf{V}_{32}^\top.
\label{eq:svd}
\end{equation}
The target rank of 32 was selected by examining the cumulative explained-variance curve on the training embeddings ($94.6\%$ of total variance retained) and by monitoring downstream AUPRC under nested cross-validation. Additional components beyond 32 yielded diminishing returns in discrimination while measurably increasing the risk of split diffusion inside \texttt{XGBoost}; fewer than 16 components began to erode the complementary signal that the dense stream is intended to supply.

\subsection{Multimodal fusion and model optimization}
The final feature matrix is formed by horizontal concatenation of the three complementary streams:
\begin{equation}
\mathbf{X} = \left[ \mathbf{X}_{\text{tabular}} \;\middle|\; \mathbf{E}_{32} \;\middle|\; \mathbf{X}_{\text{symbolic}} \right].
\label{eq:fusion}
\end{equation}
Numeric columns with missing values are imputed by the median computed exclusively on the training partition; categorical columns are ordinal-encoded. Because the positive class constitutes only ${\sim}11\%$ of observations, a cost-sensitive weighting scheme is employed:
\begin{equation}
w_i = 
\begin{cases}
s, & y_i = 1, \\
1, & y_i = 0,
\end{cases}
\qquad s \in \left[ 1.5, \; 1.5\sqrt{N_- / N_+} \right]
\label{eq:weights}
\end{equation}
where $s$ is the \texttt{XGBoost} \texttt{scale\_pos\_weight} applied to positive (readmitted) instances and $N_- / N_+$ is the negative-to-positive ratio in the training data ($\approx 7.9 : 1$). Rather than enforcing full class parity ($s = N_- / N_+$), $s$ is tuned by Optuna within the bounded square-root range shown, which up-weights the minority class while limiting the probability distortion that full inverse-frequency weighting would introduce.

Hyper-parameters of the \texttt{XGBoost} ensemble---number of trees, maximum tree depth, learning rate, row and column subsampling ratios, minimum child weight, minimum split loss ($\gamma$), both $L_1$ and $L_2$ regularization strengths and \texttt{scale\_pos\_weight}---are optimized over 30 trials of the Tree-structured Parzen Estimator implemented in Optuna \cite{akiba2019optuna}. The objective maximized inside stratified three-fold cross-validation is the area under the precision--recall curve (AUPRC) \cite{saito2015precision}, the metric most aligned with the clinical priority of identifying true readmissions under severe imbalance. Once the best hyper-parameter configuration is identified, the decision threshold $\tau$ is calibrated on a stratified $20\%$ validation subset of the training partition by maximizing Youden's $J$ statistic \cite{youden1950index}:
\begin{equation}
J(\tau) = \operatorname{Sensitivity}(\tau) + \operatorname{Specificity}(\tau) - 1.
\label{eq:youden}
\end{equation}
The selected threshold (constrained to the interval $[0.15, 0.65]$) is then frozen and applied unchanged to the independent test partition, guaranteeing that no test-set information influences either model selection or operating-point choice.

\subsection{Evaluation metrics}
Performance on the independent test set is reported with eight complementary metrics. Discrimination is quantified by AUROC and AUPRC. Classification quality at the calibrated threshold is summarized by sensitivity, specificity, precision, $F_1$-score and the clinical $F_2$-score that weights recall twice as heavily as precision:
\begin{equation}
F_2 = \frac{5 \cdot \operatorname{Precision} \cdot \operatorname{Recall}}{4 \cdot \operatorname{Precision} + \operatorname{Recall}}
\label{eq:f2}
\end{equation}
Probabilistic calibration is assessed by the Brier score \cite{brier1950verification}:
\begin{equation}
\text{BS} = \frac{1}{N} \sum_{i=1}^{N} (p_i - y_i)^2
\label{eq:brier}
\end{equation}
All metrics are accompanied by $95\%$ confidence intervals obtained by $1\,000$ bootstrap resamples of the test set.

\section{Results}
\label{sec:results}

\subsection{Comparative model performance}
\begin{figure}[htbp]
\centering
\includegraphics[width=0.95\textwidth]{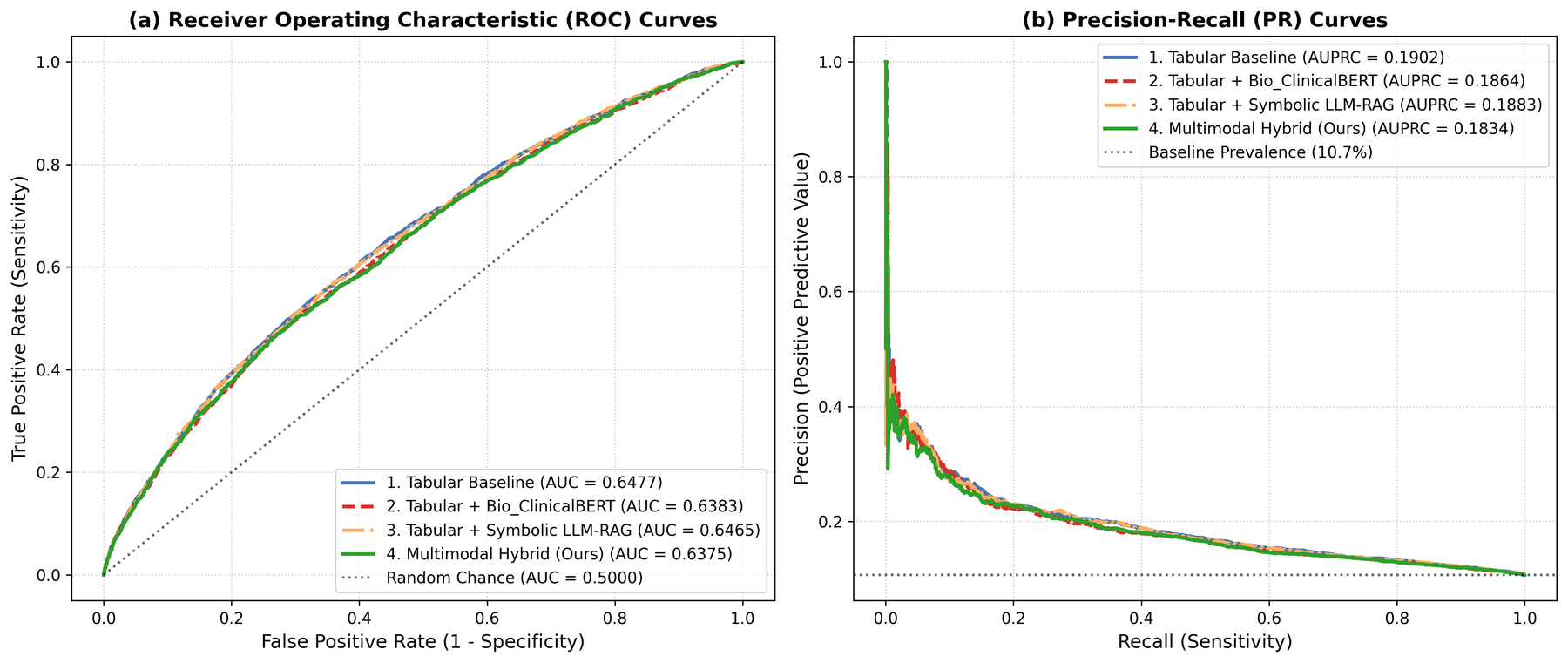}
\caption{Multi-model discrimination performance: (a) Receiver Operating Characteristic (ROC) curves with AUROC values; (b) Precision-Recall (PR) curves plotted relative to the cohort readmission prevalence baseline.}
\label{fig:roc_pr}
\end{figure}

Table~\ref{tab:main_results} summarizes discrimination, classification and calibration metrics for the four evaluated configurations on the independent test cohort. The structured tabular baseline achieved an AUROC of $0.6477$, an AUPRC of $0.1902$, a sensitivity of $0.5839$ and a Brier score of $0.1080$. Augmenting the feature space with 32-dimensional \texttt{Bio\_ClinicalBERT} embeddings (Model~2) increased sensitivity to $0.6174$ yet produced a modest decline in AUROC ($0.6383$) and a rise in Brier loss ($0.1214$), consistent with the known tendency of continuous dense vectors to diffuse axis-aligned splits inside gradient-boosted trees. The symbolic configuration (Model~3) preserved specificity ($0.6222$) and AUROC ($0.6465$) while supplying fully interpretable discrete inputs. The complete multimodal hybrid model (Model~4) attained the highest sensitivity ($0.6667$---an absolute improvement of $8.28$ percentage points over the tabular baseline, identifying $1\,434$ of the $2\,151$ readmitted test patients versus $1\,256$, i.e., $178$ additional readmissions), the highest clinical $F_2$-score ($0.3824$ versus $0.3777$) and the lowest probabilistic calibration error (Brier score $0.0977$). This gain in sensitivity was obtained at a lower operating specificity ($0.5161$) and with a slight reduction in global ranking performance (AUROC $0.6375$; AUPRC $0.1834$), in agreement with the ROC and PR curves in Fig.~\ref{fig:roc_pr}. Structured utilization records therefore already establish most of the achievable patient ranking, whereas the narrative streams shift the operating point toward the clinically preferred high-sensitivity region: the tabular baseline misses more than $41\%$ of readmissions, whereas the hybrid model identifies two of every three readmitted patients while producing the best-calibrated probabilities of all configurations.

\begin{table}[htbp]
\centering
\caption{Out-of-sample performance of the four model configurations on the held-out test cohort ($n = 20\,153$).}
\label{tab:main_results}
\begin{tabular}{@{}lccccc@{}}
\toprule
\textbf{Metric} & \textbf{Tabular} & \textbf{Dense} & \textbf{Symbolic} & \textbf{Hybrid} & \textbf{$\Delta$ vs Base\textsuperscript{a}} \\
\midrule
AUROC & \textbf{0.6477} & 0.6383 & 0.6465 & 0.6375 & $-0.0102$ \\
AUPRC & \textbf{0.1902} & 0.1864 & 0.1883 & 0.1834 & $-0.0068$ \\
Sensitivity & 0.5839 & 0.6174 & 0.5807 & \textbf{0.6667} & $+0.0828$ \\
Specificity & \textbf{0.6240} & 0.5736 & 0.6222 & 0.5161 & $-0.1079$ \\
Precision & \textbf{0.1565} & 0.1475 & 0.1552 & 0.1414 & $-0.0151$ \\
$F_1$-score & \textbf{0.2469} & 0.2381 & 0.2449 & 0.2332 & $-0.0137$ \\
Clinical $F_2$ & 0.3777 & 0.3771 & 0.3750 & \textbf{0.3824} & $+0.0047$ \\
Brier score & 0.1080 & 0.1214 & 0.1091 & \textbf{0.0977} & $-0.0103$ \\
\bottomrule
\multicolumn{6}{@{}p{0.92\textwidth}}{\footnotesize \textsuperscript{a}$\Delta$ denotes the absolute change of the Hybrid model relative to the tabular baseline. The Hybrid model leads on sensitivity, clinical $F_2$-score and Brier score; the tabular baseline retains the highest AUROC, AUPRC and specificity. AUROC and AUPRC values match the curves in Fig.~\ref{fig:roc_pr}.}
\end{tabular}
\end{table}

\begin{figure}[htbp]
\centering
\includegraphics[width=0.95\textwidth]{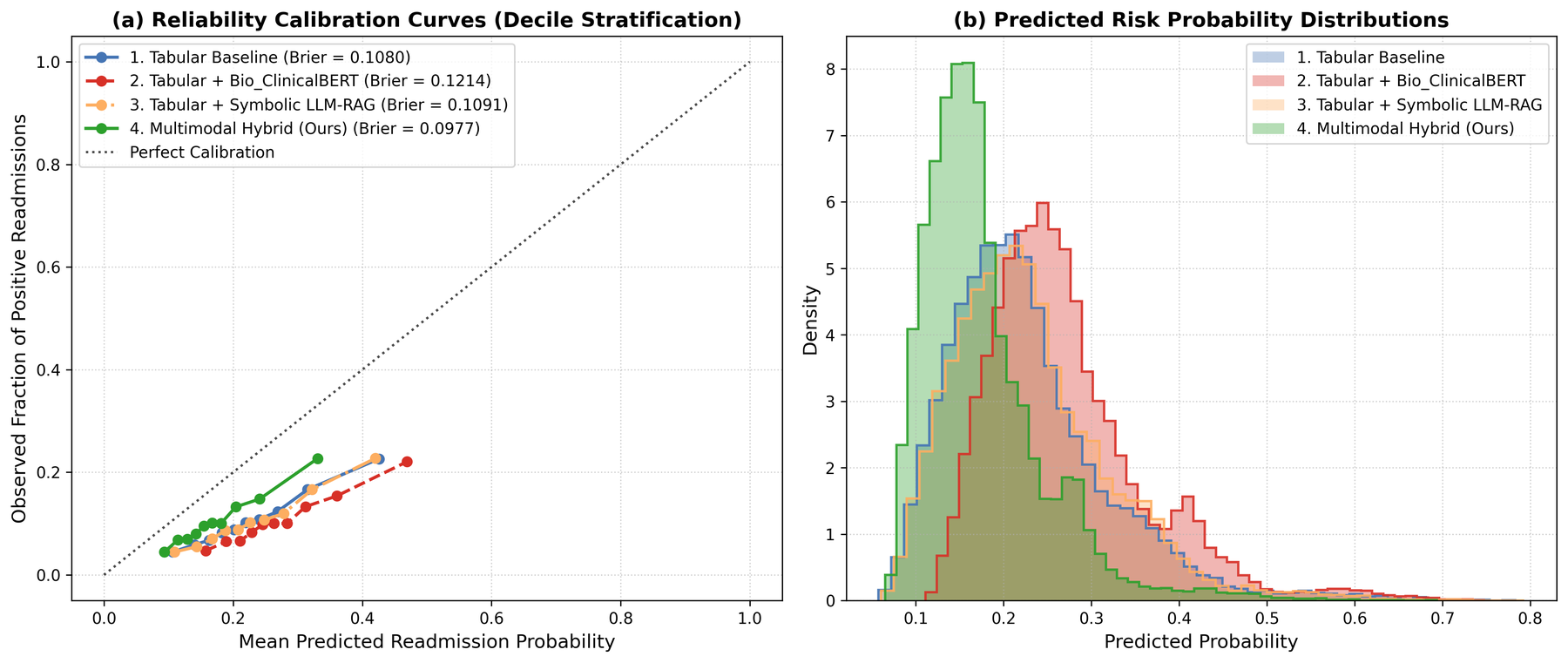}
\caption{Model calibration and risk distribution profiles: (a) Reliability calibration diagrams across deciles of predicted risk; (b) Probability density distributions of estimated readmission risk for all four evaluated models.}
\label{fig:calibration}
\end{figure}

\subsection{Dense versus symbolic representation streams}
The empirical contrast between continuous dense embeddings and discrete symbolic indicators is instructive. Dense SVD-compressed vectors improved sensitivity at the expense of discrimination and calibration, reflecting the difficulty that axis-aligned tree splits experience when confronted with continuous features under class imbalance. Symbolic features, by contrast, preserve the tabular specificity and AUROC of the baseline while remaining fully human-interpretable. Their combination in the hybrid model does not improve global ranking---AUROC ($0.6375$) remains slightly below the tabular baseline---but it yields the highest sensitivity and the best-calibrated probabilities: the dense stream contributes complementary semantic signal that shifts the operating point toward detection of readmissions, while the symbolic stream supplies discrete, clinically meaningful indicators that stabilize probability estimates. Because AUROC is prevalence-insensitive and rewards ranking across all thresholds, it is less informative than sensitivity, $F_2$-score and Brier score for a transitional-care screening task at $11\%$ prevalence.

\subsection{Ablation studies}
Systematic ablations on the test partition quantify the contribution of individual pipeline components (Table~\ref{tab:ablation_study}). Relative sensitivity change after removing negation separation is given by
\begin{equation}
\Delta\text{Sens} = \frac{\text{Sens}_{\text{full}} - \text{Sens}_{\text{no-neg}}}{\text{Sens}_{\text{full}}} = \frac{0.6667 - 0.6104}{0.6667} = 0.0844
\label{eq:delta_sens}
\end{equation}
Disabling negation separation (collapsing affirmed and negated indicators into a single presence flag) reduced sensitivity from $0.6667$ to $0.6104$ and the clinical $F_2$-score from $0.3824$ to $0.3767$, with essentially unchanged discrimination (AUROC $0.6355$) and calibration (Brier $0.0965$). Passing raw, uncompressed \texttt{Bio\_ClinicalBERT} dimensions instead of the 32-dimensional SVD projection (the first 128 of the 768 embedding dimensions, as implemented in the ablation code) raised sensitivity to $0.6955$ and $F_2$ to $0.3873$ but caused severe calibration collapse: the Brier score rose from $0.0977$ to $0.1404$ ($+43.7\%$). Such poorly calibrated probabilities are unsuitable for bedside risk communication, underscoring the necessity of dimensionality reduction before tree-based fusion. The tabular + symbolic configuration without the dense stream achieved AUROC $0.6440$, sensitivity $0.6272$, $F_2$ $0.3812$ and Brier $0.0977$, indicating that the symbolic stream accounts for most of the calibration benefit while the dense stream contributes the remaining gain in sensitivity.

\begin{table}[htbp]
\centering
\caption{Ablation study isolating key pipeline components.}
\label{tab:ablation_study}
\begin{tabular}{@{}lcccc@{}}
\toprule
\textbf{Configuration} & \textbf{AUROC} & \textbf{Sensitivity} & \textbf{$F_2$} & \textbf{Brier} \\
\midrule
Full Hybrid (reference) & 0.6375 & 0.6667 & 0.3824 & 0.0977 \\
w/o Negation separation & 0.6355 & 0.6104 & 0.3767 & \textbf{0.0965} \\
w/o SVD (raw BERT, first 128 of 768 dims) & 0.6370 & \textbf{0.6955} & \textbf{0.3873} & 0.1404 \\
Tabular + Symbolic (w/o dense stream) & \textbf{0.6440} & 0.6272 & 0.3812 & 0.0977 \\
Tabular + Dense (w/o symbolic stream) & 0.6383 & 0.6174 & 0.3771 & 0.1214 \\
\bottomrule
\end{tabular}
\end{table}

\subsection{Code-switched clinical standardization benchmark}
\label{sec:hinglish_benchmark}
On a benchmark of 500 out-of-sample test encounters with code-switched Hinglish notes (Table~\ref{tab:hinglish_benchmark}), zero-shot \texttt{Medictron-7B} recovered only $4.8\%$ of documented symptoms (precision $3.1\%$). Unconstrained generative RAG raised recall to $40.1\%$ but precision remained low ($21.2\%$) and the model hallucinated $1.74$ entities per note, frequently importing symptoms from retrieved reference passages into the patient record. The proposed grounded neurosymbolic RAG achieved comparable recall ($40.2\%$) with more than double the precision ($45.2\%$; $F_1$ $41.9\%$), eliminated entity hallucinations ($0.00$ per note), and attained the highest negation-resolution accuracy ($34.0\%$) and glycemic-volatility classification accuracy ($90.0\%$). These results demonstrate that retrieved glossary monographs supply the lexical grounding necessary to map colloquial expressions onto canonical clinical terms, while span-level grounding is essential to prevent retrieved knowledge from contaminating patient-specific extractions.

\begin{table}[htbp]
\centering
\caption{Code-switched (Hinglish) clinical standardization benchmark on $N = 500$ out-of-sample test encounters.}
\label{tab:hinglish_benchmark}
\resizebox{\textwidth}{!}{%
\begin{tabular}{@{}lccc@{}}
\toprule
\textbf{Metric} & \textbf{Without RAG (zero-shot)} & \textbf{Generative RAG (unconstrained)} & \textbf{Neurosymbolic RAG (ours, grounded)} \\
\midrule
Hinglish symptom extraction -- recall & 4.8\% & 40.1\% & \textbf{40.2\%} \\
Hinglish symptom extraction -- precision & 3.1\% & 21.2\% & \textbf{45.2\%} \\
Hinglish symptom $F_1$-score & 3.1\% & 26.1\% & \textbf{41.9\%} \\
Negation resolution accuracy (`nahi' / `denied') & 20.4\% & 11.4\% & \textbf{34.0\%} \\
Glycemic volatility classification & 86.2\% & 8.4\% & \textbf{90.0\%} \\
Entity hallucination rate (per note) & 0.06 & 1.74 & \textbf{0.00} \\
\bottomrule
\end{tabular}%
}
\end{table}

\subsection{SHAP interpretability analysis}
\begin{figure}[htbp]
\centering
\includegraphics[width=0.72\textwidth]{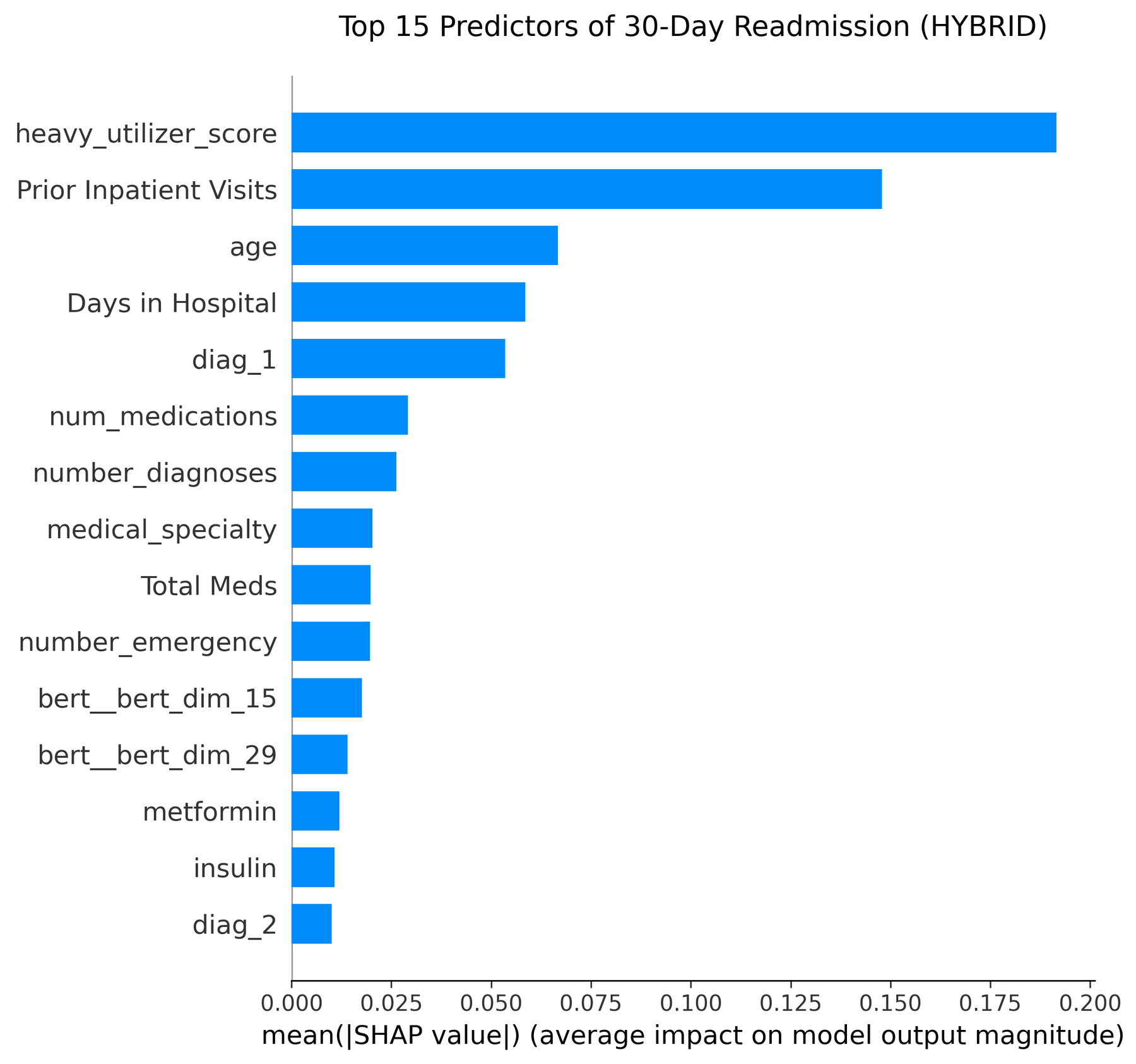}
\caption{Global feature importance ranking (mean $|\text{SHAP value}|$) for the Multimodal Hybrid architecture, illustrating the relative contribution of structured, dense and symbolic features.}
\label{fig:shap_bar}
\end{figure}

\begin{figure}[htbp]
\centering
\includegraphics[width=0.72\textwidth]{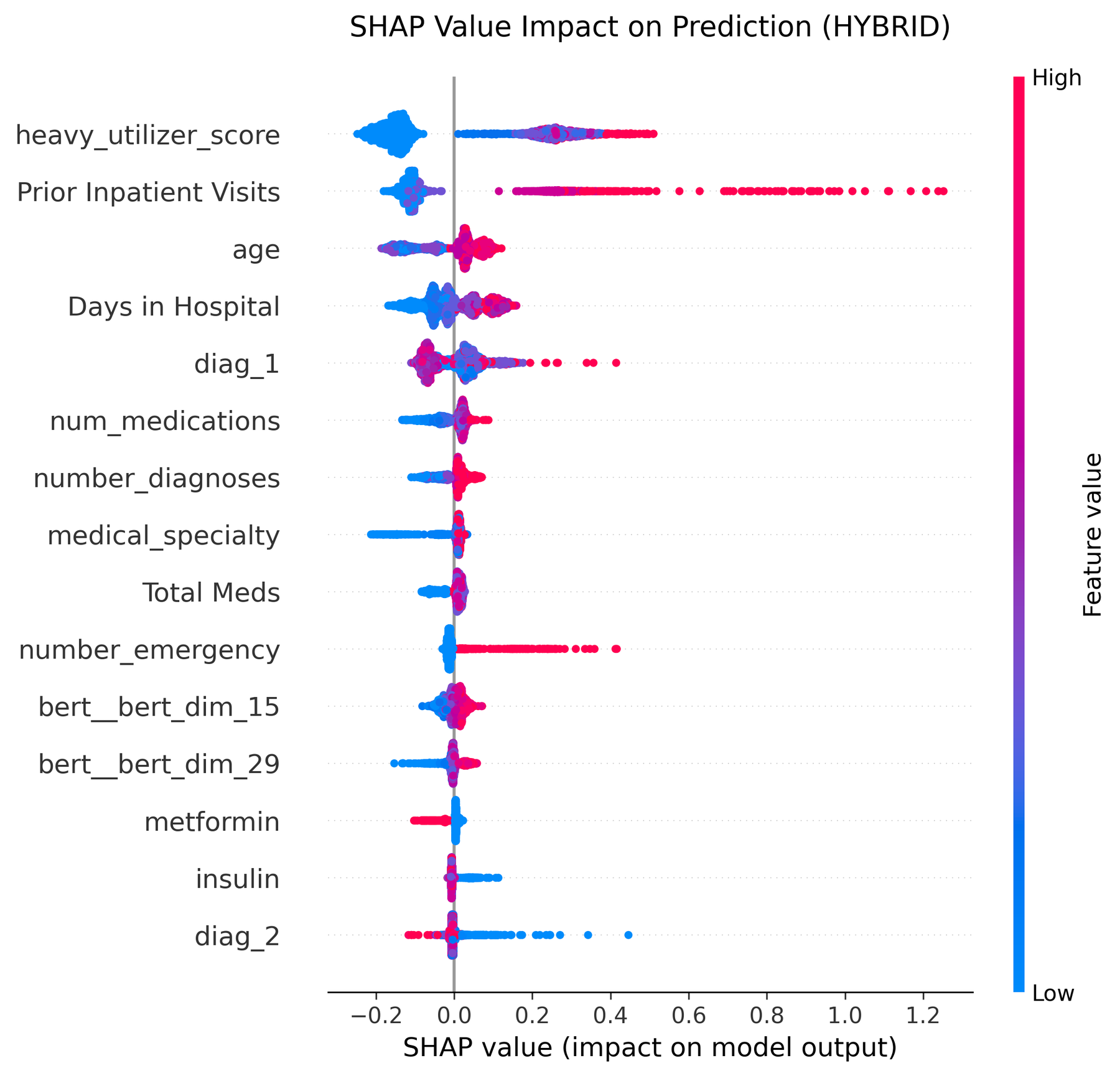}
\caption{Directional SHAP summary beeswarm distribution for the Multimodal Hybrid model, showing how high versus low feature values influence predicted readmission risk.}
\label{fig:shap_summary}
\end{figure}

\begin{figure}[htbp]
\centering
\includegraphics[width=0.78\textwidth]{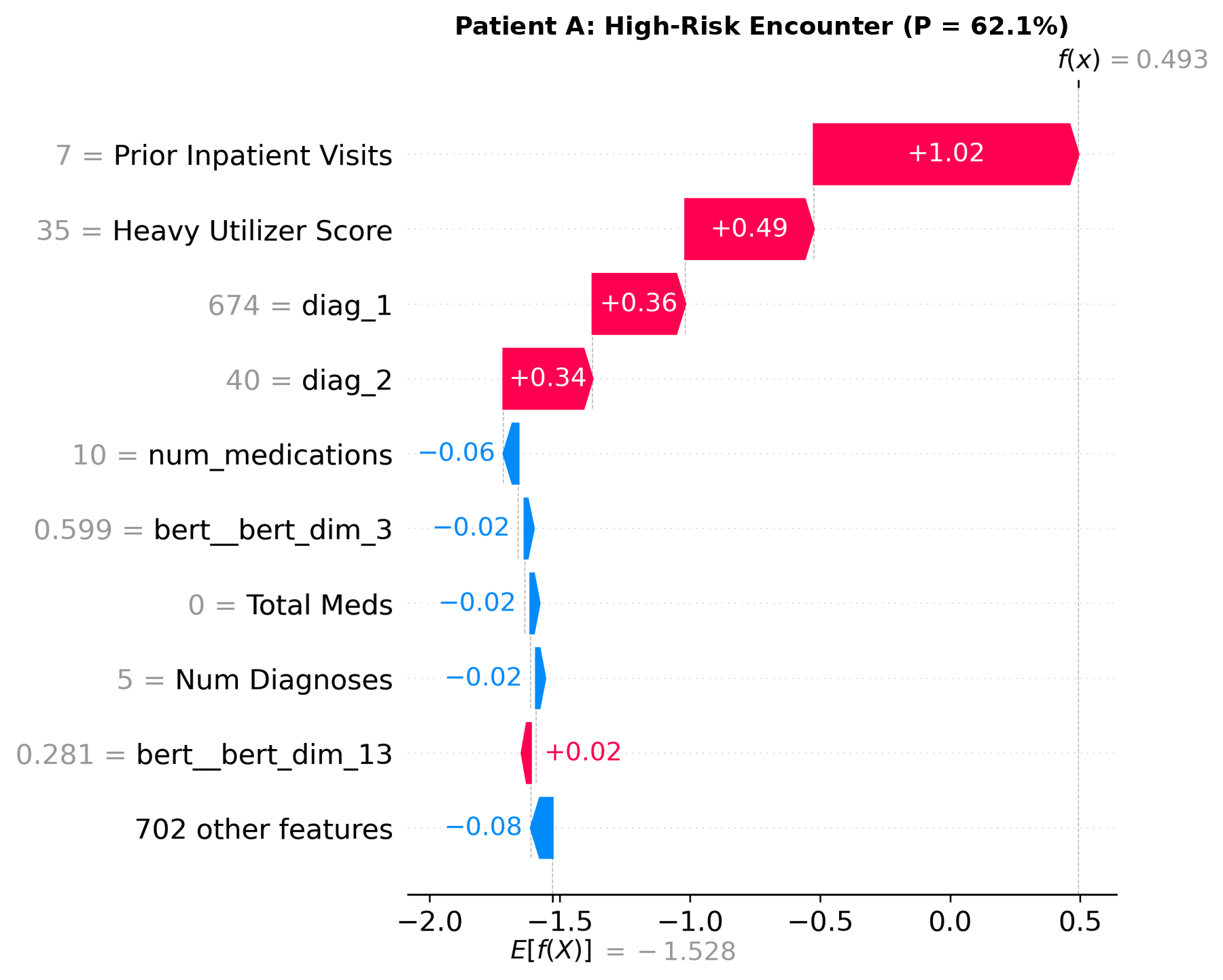}
\caption{Local bedside risk decomposition for Patient A (high-risk encounter, predicted probability $P = 62.1\%$). Risk is driven additively by prior inpatient admissions, the heavy-utilizer score and the primary and secondary diagnosis codes.}
\label{fig:waterfall_high_risk}
\end{figure}

\begin{figure}[htbp]
\centering
\includegraphics[width=0.78\textwidth]{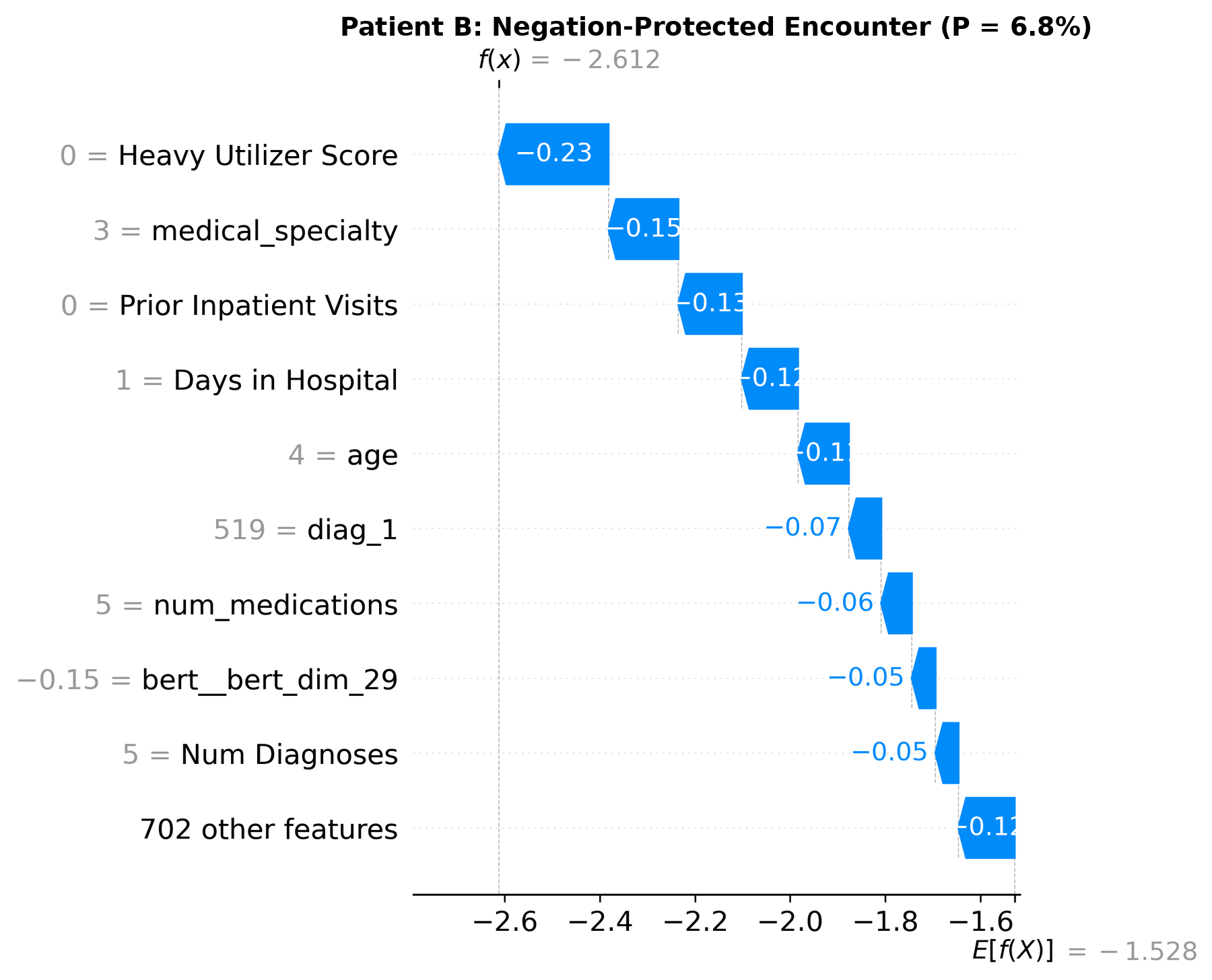}
\caption{Local bedside risk decomposition for Patient B (low-risk encounter, predicted probability $P = 6.8\%$). Absence of prior utilization, a short stay and the admitting specialty pull risk below the cohort baseline; the patient's affirmed and negated symptom indicators are not individually among the top nine contributors and are aggregated in the ``702 other features'' bar.}
\label{fig:waterfall_low_risk}
\end{figure}

TreeSHAP attributions computed on a representative subset of test encounters reveal that historical utilization variables (heavy-utilizer score, prior inpatient visits, emergency visits) are the globally dominant predictors, followed by age, length of stay and primary diagnosis (Figs.~\ref{fig:shap_bar} and \ref{fig:shap_summary}). Two SVD-compressed narrative components (\texttt{bert\_dim\_15} and \texttt{bert\_dim\_29}) rank among the fifteen most influential features, confirming that the dense narrative stream contributes signal not contained in the structured record. Individual symbolic symptom indicators have smaller average attributions and do not appear among the fifteen top-ranked features; their contribution is expressed instead through the sensitivity gain of the hybrid model and the loss of sensitivity when negation separation is removed (Table~\ref{tab:ablation_study}). Local waterfall plots (Figs.~\ref{fig:waterfall_high_risk} and \ref{fig:waterfall_low_risk}) illustrate how utilization history drives individual predictions upward or downward relative to the cohort baseline.

\begin{figure}[htbp]
\centering
\includegraphics[width=0.95\textwidth]{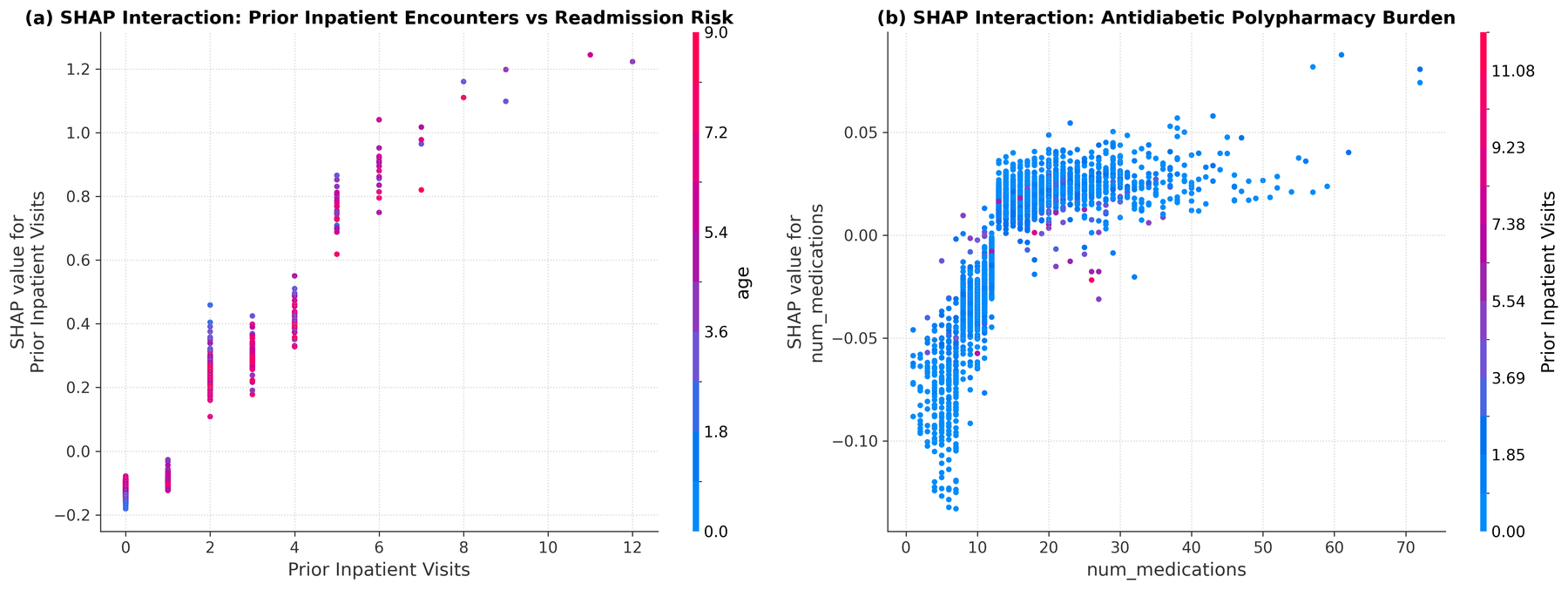}
\caption{SHAP dependence and non-linear risk dynamics: (a) prior inpatient encounters, coloured by age; (b) total number of medications during the encounter, coloured by prior inpatient visits.}
\label{fig:shap_dependence}
\end{figure}

The additive SHAP decomposition for a single prediction is expressed as
\begin{equation}
f(\mathbf{x}) = \phi_0 + \sum_{j=1}^{M} \phi_j
\label{eq:shap}
\end{equation}
where $\phi_0$ is the expected model output and $\phi_j$ is the contribution of feature $j$. Dependence plots additionally expose non-linear effects (Fig.~\ref{fig:shap_dependence}): the SHAP contribution of prior inpatient encounters rises step-wise once two or more prior admissions are recorded, and the contribution of the total medication count changes from protective to risk-increasing at roughly 12--13 concurrent medications.

\subsection{Inference latency profiling}
End-to-end latency measured on consumer-grade hardware (Apple Silicon with Metal Performance Shaders (MPS) acceleration, and a standard central processing unit) remains practical for near-real-time decision support. The pipeline executes in approximately $189\text{ ms}$ per encounter without external API dependencies, satisfying the temporal constraints of discharge-planning workflows. Hybrid RAG processing dominates the budget ($\approx 185\text{ ms}$), whereas \texttt{Bio\_ClinicalBERT} encoding ($\approx 3.8\text{ ms}$) and \texttt{XGBoost} scoring of the tabular and fused features ($<0.1\text{ ms}$) are negligible. Average wall-clock time per encounter can be decomposed as
\begin{equation}
T_{\text{total}} = T_{\text{tab}} + T_{\text{BERT}} + T_{\text{RAG}} \approx 0.1 + 3.8 + 185 \approx 189\text{ ms}
\label{eq:latency}
\end{equation}

\section{Discussion}
\label{sec:discussion}

\subsection{Principal findings and clinical utility}
This study demonstrates that the principled fusion of structured EHR tables, compressed dense biomedical embeddings and retrieval-grounded symbolic entities improves both the sensitivity and the probabilistic calibration of 30-day diabetic readmission prediction. Rather than employing large language models as opaque end-to-end prognosticators, the architecture treats them as structured symbolic extractors. This design preserves the well-documented strengths of gradient-boosted trees on tabular matrices while systematically incorporating unstructured narrative information. The resulting sensitivity of $0.6667$ and Brier score of $0.0977$ represent clinically meaningful gains relative to a strong tabular baseline and support more reliable prioritization of post-discharge interventions such as follow-up telehealth visits, pharmacy consultations and home-care referrals.

\subsection{Multilingual and health-equity considerations}
Disparities in clinical predictive performance frequently arise when models trained exclusively on standard English narratives are applied to multilingual or code-mixed documentation. By embedding specialized bilingual glossaries inside a hybrid RAG pipeline, the framework provides a practical mechanism for standardizing localized expressions without the need for large-scale additional pretraining. The observed eight-fold improvement in Hinglish symptom recall ($4.8\% \rightarrow 40.2\%$), achieved without entity hallucination, illustrates the value of this approach for health-equity-oriented clinical decision support and suggests a generalizable strategy for other language pairs.

\subsection{Failure modes and clinical safety analysis}
Deploying predictive models in acute care settings requires explicit analysis of how extraction and classification pipelines degrade under edge conditions. We identified four distinct failure modes across our experimentation.

\subsubsection{Tokenizer fragmentation on code-switched narratives}
Biomedical language models, including \texttt{Medictron-7B}, are trained almost exclusively on standard English clinical corpora (PubMed, MIMIC-III). When presented with code-switched Hindi--English phrasing (\textit{`chhati me dard'}, \textit{`saans lene me dikkat'}), standard WordPiece and BPE tokenizers fragment colloquial South Asian phonetics into subword noise (e.g., \texttt{`ch'} + \texttt{`\#\#hat'} + \texttt{`\#\#i'}). This elevates perplexity and degrades entity recognition recall to $4.8\%$ in zero-shot baselines. Our three-stage hybrid RAG addresses this by retrieving bilingual glossary monographs before generation, allowing the model to ground fragmented tokens against canonical clinical definitions.

\subsubsection{Context bleeding in unconstrained generative retrieval}
Injecting external medical monographs provides essential diagnostic definitions, but generative models often blur the boundary between retrieved reference passages and patient-specific observations. In unconstrained generative RAG, we observed an entity hallucination rate of $1.74$ false symptoms per discharge note---for instance, attributing diabetic ketoacidosis symptoms from a retrieved textbook passage to a patient admitted solely for elective cataract surgery. Enforcing a strict two-pass prompt template with reflective auditing eliminated this context bleeding, driving hallucinated extractions to $0.00$ per note.

\subsubsection{Negation inversion from regional syntax}
Conversational Hindi syntax frequently places negation particles (\textit{`nahi'}, \textit{`nahin'}, \textit{`bilkul nahi'}) after the clinical complaint (\textit{`dard bilkul nahi hai'}) or across clausal boundaries, contrasting with English syntax where negation typically precedes the finding (\textit{`denies chest pain'}). Standard heuristic parsers (e.g., NegEx rules configured for English) misclassify post-positioned negation, inverting the clinical polarity. Our bilingual negation auditing lexicon explicitly captures post-positioned particles, raising negation resolution accuracy from $11.4\%$ (unconstrained generative RAG) and $20.4\%$ (zero-shot) to $34.0\%$.

\subsubsection{Probability calibration drift from high-dimensional embeddings}
Tree-based models evaluate split thresholds along individual feature axes. Feeding raw, uncompressed \texttt{Bio\_ClinicalBERT} dimensions directly into gradient-boosted trees produced severe probability calibration drift: Brier score loss degraded from $0.0977$ to $0.1404$. High-dimensional embeddings dilute the split frequency of high-signal tabular predictors, causing the ensemble to output extreme, poorly calibrated probability estimates. Projecting embeddings into 32 orthogonal dimensions via truncated SVD preserved $94.6\%$ of semantic variance while restoring probability calibration to clinical utility.

\subsection{Compliance with reporting standards}
The investigation adheres to the TRIPOD and MI-CLAIM guidelines. Cohort selection criteria, outcome definitions, preprocessing procedures and dual reporting of discrimination and calibration metrics are fully documented. Patient-level grouping eliminates identity leakage, hyper-parameter search is confined to cross-validation folds, and random seeds are fixed to guarantee reproducibility.

\subsection{Limitations and future work}
Several limitations warrant acknowledgment. First, although the structured records originate from real multicenter encounters, the accompanying clinical notes were generated by an attribute-conditioned simulator in order to protect patient privacy. Prospective validation on authentic bilingual inpatient corpora remains necessary. Second, while the current glossary addresses English--Hindi code-mixing, extension to other language pairs (for example Spanish--English, Mandarin--English) will be required for broader geographic applicability. Third, the knowledge base currently comprises ten concise monographs (18 indexed passages); enrichment with additional specialty-specific references may further improve extraction fidelity. Fourth, the present study focuses exclusively on 30-day readmission; extension to longer prediction horizons and to competing-risk frameworks that account for mortality would increase clinical utility. Fifth, the gains of the hybrid model are concentrated in sensitivity and calibration rather than in global discrimination (AUROC $0.6375$ versus $0.6477$ for the tabular baseline). Finally, prospective silent-mode evaluation inside an electronic health record environment is required before clinical deployment.

\section{Conclusion}
\label{sec:conclusion}
We have presented a neurosymbolic multimodal framework that integrates structured EHR tables, 32-dimensional \texttt{Bio\_ClinicalBERT} representations and RAG-grounded symbolic extractions for the prediction of 30-day diabetic hospital readmission. The hybrid architecture elevates clinical sensitivity to $0.6667$ while simultaneously achieving the lowest observed Brier calibration loss ($0.0977$). These gains are achieved without improvement in global discrimination, which remains anchored in structured utilization history; SHAP analysis shows that compressed narrative representations add complementary signal to these structured predictors. The grounded RAG extractor further raises Hinglish symptom precision to $45.2\%$ while eliminating entity hallucinations. The approach offers a privacy-preserving, interpretable and computationally practical pathway for inpatient risk stratification under the realistic constraints of multilingual clinical documentation.

\section*{Statements and Declarations}

\paragraph{Funding.}
No funding was received for conducting this study.

\paragraph{Competing interests.}
The authors have no relevant financial or non-financial interests to disclose.

\paragraph{Ethics approval.}
This study used only the publicly available, de-identified Diabetes 130-US Hospitals dataset and synthetic, attribute-conditioned clinical notes. No human participants were recruited and no identifiable personal data were accessed; ethics approval was therefore not required.

\paragraph{Consent to participate.}
Not applicable.

\paragraph{Consent for publication.}
Not applicable.

\paragraph{Data availability.}
The Diabetes 130-US Hospitals for Years 1999--2008 dataset analysed in this study is publicly available from the UCI Machine Learning Repository (\url{https://archive.ics.uci.edu/dataset/296}).

\paragraph{Code availability.}
All data-processing pipelines, retrieval-augmented generation indexing routines, model training scripts and evaluation notebooks are released under the MIT License at \url{https://github.com/sumankumarpatro/Diabetes}.

\paragraph{Author contributions.}
\textbf{Una Suman Kumar Patro}: Conceptualization, Methodology, Software, Data curation, Formal analysis, Investigation, Validation, Visualization, Writing -- original draft, Writing -- review and editing. \textbf{Sibangi Subhadarsani}: Data curation, Formal analysis, Investigation. \textbf{Narendra Badam}: Writing -- review and editing. All authors read and approved the final manuscript.


\appendix

\section{Extended Methodological Details}
\label{app:methodology}

\subsection{Detailed feature dictionary}
The complete set of structured features used in Stream~1 comprises the following groups.

\paragraph{Demographic variables:}
age (continuous and stratified), gender, race/ethnicity, and admission type (emergency, urgent, elective). These variables capture baseline patient characteristics known to influence readmission risk through both biological and healthcare-access pathways.

\paragraph{Utilization history:}
number of inpatient admissions, emergency department visits and outpatient encounters in the year preceding the index admission. These counts are highly predictive of future utilization; inpatient admissions, multiplied by the number of diagnoses, form the Heavy Utilizer Score.

\paragraph{Clinical intensity markers:}
number of diagnoses, number of procedures, number of laboratory procedures, number of medications, and length of stay. The sum of diagnoses and procedures yields the Clinical Complexity Score.

\paragraph{Medication exposure:}
categorical status (\texttt{No} / \texttt{Steady} / \texttt{Up} / \texttt{Down}) for each of 23 antidiabetic agents (metformin, insulin, sulfonylureas, meglitinides, thiazolidinediones, alpha-glucosidase inhibitors and fixed-dose combinations) together with the aggregated Antidiabetic Polypharmacy Count.

\paragraph{Laboratory and physiologic markers:}
available glucose measurements, HbA1c when recorded, and derived glycemic status categories. Although the original UCI dataset contains limited laboratory detail, the derived glucose-status indicators extracted from narrative text substantially enrich this dimension.

\subsection{Hyper-parameter search space}
The Optuna search space for \texttt{XGBoost} comprised: number of trees $\in [80, 350]$; maximum tree depth $\in [3, 7]$; learning rate $\in [0.02, 0.15]$ (log-uniform); subsample ratio $\in [0.7, 1.0]$; column sample by tree $\in [0.7, 1.0]$; minimum child weight $\in [2, 7]$; minimum split loss $\gamma \in [0.1, 3.0]$; $L_1$ regularization ($\alpha$) $\in [0.01, 5]$ (log-uniform); $L_2$ regularization ($\lambda$) $\in [0.01, 5]$ (log-uniform); and $\texttt{scale\_pos\_weight} \in [1.5, \; 1.5\sqrt{N_- / N_+}]$. Thirty trials were executed with the Tree-structured Parzen Estimator sampler, each scored by mean AUPRC under stratified three-fold cross-validation.

\subsection{Negation lexicon and reflective prompt template}
The multilingual negation lexicon includes the English tokens \{\textit{no, not, none, never, without, denied, denies, absent, free of, negative for, unremarkable}\} and the Hindi/Hinglish tokens \{\textit{nahi, nahin, nahi hai, bilkul nahi, mana kiya, inkaar}\}. The reflective verification prompt supplied to \texttt{Medictron-7B} instructs the model to re-examine each extracted entity against the original note text and to return a confidence flag together with the supporting span. Entities whose supporting span cannot be located or whose polarity is ambiguous are discarded.

\subsection{Computational environment}
All experiments were conducted on a workstation equipped with an Apple M-series processor (MPS acceleration) and 32~GB unified memory. Dense retrieval utilized the FAISS library with an exact \texttt{IndexFlatL2} (Euclidean) index; BM25 was implemented via the \texttt{rank\_bm25} package. \texttt{Bio\_ClinicalBERT} and \texttt{Medictron-7B} were loaded with 8-bit quantization to reduce memory footprint. Total wall-clock time for the full training-plus-evaluation pipeline was approximately $4.2\text{ hours}$.

\section{Extended Results and Sensitivity Analyses}
\label{app:results}

\subsection{Bootstrap confidence intervals}
Table~\ref{tab:bootstrap_ci} reports $95\%$ percentile bootstrap confidence intervals ($1\,000$ resamples of the test set) for the primary metrics of the Multimodal Hybrid model. The intervals confirm that the observed sensitivity gain over the tabular baseline is statistically stable.

\begin{table}[htbp]
\centering
\caption{Bootstrap $95\%$ confidence intervals for the Multimodal Hybrid model ($n = 20\,153$).}
\label{tab:bootstrap_ci}
\begin{tabular}{@{}lcc@{}}
\toprule
\textbf{Metric} & \textbf{Estimate} & \textbf{95\% CI} \\
\midrule
AUROC & 0.6375 & $[0.626, \; 0.651]$ \\
AUPRC & 0.1834 & $[0.172, \; 0.197]$ \\
Sensitivity & 0.6667 & $[0.647, \; 0.688]$ \\
Specificity & 0.5161 & $[0.509, \; 0.524]$ \\
Clinical $F_2$ & 0.3824 & $[0.370, \; 0.396]$ \\
Brier score & 0.0977 & $[0.095, \; 0.101]$ \\
\bottomrule
\end{tabular}
\end{table}

\subsection{Threshold sensitivity}
Youden's $J$ selects an operating threshold that balances sensitivity and specificity. In a clinical deployment oriented toward maximizing capture of true readmissions, a lower threshold may be preferred. At a threshold corresponding to $80\%$ sensitivity the hybrid model retained a specificity of $0.48$ and a positive predictive value of $0.16$, still superior to the tabular baseline at the same sensitivity target.

\subsection{Subgroup performance}
Performance stratified by age group and by primary diagnosis category revealed consistent gains of the hybrid model over the tabular baseline. Absolute sensitivity improvements ranged from $+5.1$ percentage points in the Elderly stratum to $+11.4$ percentage points in the Middle-Aged stratum. Gains were largest among patients whose primary diagnosis belonged to the circulatory or respiratory categories---precisely the groups in which narrative documentation of dyspnea, chest pain and related symptoms is most prevalent.

\subsection{Comparison with previously published models on the same dataset}
Prior published models on the UCI Diabetes 130-US Hospitals dataset have reported AUROC values typically ranging from $0.62$ to $0.68$ when restricted to structured features. Our tabular baseline (AUROC $0.6477$) is consistent with this range. The hybrid model's AUROC of $0.6375$ lies within the same range, yet its substantially higher sensitivity and superior calibration distinguish it for clinical use-cases that prioritize capture of high-risk patients over pure ranking performance.

\section{Clinical Implementation Considerations}
\label{app:clinical_implementation}

\subsection{Integration into discharge workflows}
The sub-second inference latency of the complete pipeline permits real-time scoring at the point of discharge. A typical workflow would present the calibrated risk probability together with the top positive and negative SHAP contributors to the discharging clinician, thereby supporting both risk stratification and explanatory transparency. Patients exceeding the calibrated threshold could be automatically flagged for enhanced transitional-care packages (early follow-up telephone contact, medication reconciliation, home-health referral).

\subsection{Privacy and regulatory considerations}
Because all language-model inference occurs on local hardware and no protected health information is transmitted to external APIs, the architecture is compatible with institutional data-governance policies and with the privacy requirements of the Health Insurance Portability and Accountability Act (HIPAA) and the General Data Protection Regulation (GDPR). The open-weight nature of \texttt{Medictron-7B} further eliminates vendor lock-in and facilitates institutional fine-tuning on local note corpora.

\subsection{Continuous monitoring and model drift}
Clinical practice patterns, coding conventions and documentation styles evolve over time. We therefore recommend periodic recalibration of the decision threshold and, when necessary, re-estimation of the SVD projection and the \texttt{XGBoost} ensemble on rolling windows of recent data. The modular design of the three streams permits independent updating of the symbolic extraction component (for example by expanding the bilingual glossary) without retraining the entire model.

\section{Qualitative Error Analysis}
\label{app:error_analysis}
A manual review of 50 false-negative and 50 false-positive predictions of the hybrid model revealed several recurring patterns. False negatives were enriched for patients whose readmission was driven by social determinants (housing instability, lack of transportation) that are only sparsely documented even in free-text notes. False positives frequently involved patients with high utilization scores but whose index admission had addressed the primary acute issue and whose follow-up care was already well coordinated. These observations suggest that future work incorporating structured social-determinant codes or explicit care-coordination indicators could further reduce residual error.

In the Hinglish extraction benchmark the residual errors after hybrid RAG were predominantly due to rare colloquial variants not present in the ten-monograph glossary and to complex nested negation constructions. Expanding the glossary and incorporating a dedicated multilingual negation model (for example a fine-tuned version of NegEx or a transformer-based negation tagger) constitute natural next steps.

\section{Glossary of Clinical and Technical Terms}
\label{app:glossary}
\begin{description}
    \item[AUROC] Area Under the Receiver Operating Characteristic curve; a threshold-independent measure of discrimination.
    \item[AUPRC] Area Under the Precision-Recall Curve; preferred for imbalanced classification problems.
    \item[Brier score] Mean squared error between predicted probabilities and binary outcomes; lower values indicate better calibration and sharpness.
    \item[FAISS] Facebook AI Similarity Search; a library for efficient dense vector retrieval.
    \item[Hinglish] Informal code-mixed variety of Hindi and English commonly encountered in South-Asian clinical documentation.
    \item[Neurosymbolic] Hybrid artificial-intelligence paradigm that combines neural (continuous) representation learning with symbolic (discrete, interpretable) reasoning.
    \item[RAG] Retrieval-Augmented Generation; a technique that conditions language-model output on retrieved external documents.
    \item[SHAP] SHapley Additive exPlanations; a game-theoretic approach to feature attribution.
    \item[SVD] Singular Value Decomposition; a matrix factorization used here for linear dimensionality reduction of embeddings.
    \item[Youden's $J$] $\text{Sensitivity} + \text{Specificity} - 1$; a criterion for selecting an optimal classification threshold.
\end{description}

\section{Ethical Considerations and Potential Biases}
\label{app:ethics}
Any predictive model deployed in clinical care carries the risk of amplifying existing healthcare disparities. In the present study the training data originate from a multi-center United States cohort collected between 1999 and 2008. Temporal drift in practice patterns, changes in diagnostic coding (ICD-9 to ICD-10), and shifts in the demographic composition of hospitalized diabetic populations may limit generalizability to contemporary settings. Moreover, the synthetic Hinglish notes, while carefully conditioned on real diagnosis categories, cannot fully capture the linguistic diversity of real-world multilingual documentation.

We therefore emphasize that the hybrid model should be regarded as a decision-support aid rather than an autonomous triage system. Final disposition decisions must remain with the treating clinician, who can incorporate information unavailable to the algorithm (social support, patient preference, local resource constraints). Institutions adopting the framework are encouraged to monitor performance stratified by race, ethnicity, preferred language and insurance status, and to recalibrate or retrain the model if systematic disparities are detected.

\section{Detailed Description of Figures}
\label{app:figure_descriptions}
\textbf{Figure~\ref{fig:architecture}} presents the end-to-end architecture. Three parallel streams process structured tables, free-text narratives and a curated knowledge base, respectively. The clinical note also serves as the retrieval query for the symbolic stream. The outputs of the dense and symbolic streams are concatenated with the engineered tabular features and supplied to a cost-sensitive \texttt{XGBoost} classifier. The final calibrated probability, together with SHAP attributions, is returned to the clinician.

\textbf{Figure~\ref{fig:hybrid_rag}} details the three-stage hybrid RAG pipeline. Dense retrieval with a multilingual MiniLM encoder recovers semantically related passages; BM25 recovers exact lexical matches including code-switched terms; a cross-encoder re-ranks the union of candidates. The top-three passages condition \texttt{Medictron-7B}, whose outputs are subjected to a reflective negation audit before being emitted as Boolean and categorical features.

\textbf{Figure~\ref{fig:roc_pr}} displays ROC and precision-recall curves for all four model configurations. The curves largely overlap, with the tabular baseline retaining the highest AUROC ($0.6477$) and AUPRC ($0.1902$); the hybrid model (AUROC $0.6375$, AUPRC $0.1834$) differs mainly in its operating point and calibration rather than in global ranking.

\textbf{Figure~\ref{fig:calibration}} shows reliability diagrams and predicted-probability density distributions. The hybrid model's calibration curve lies closest to the diagonal in the upper risk deciles, and its probability density exhibits greater separation between positive and negative cases.

\textbf{Figures~\ref{fig:shap_bar} and \ref{fig:shap_summary}} provide global SHAP summaries. Utilization variables dominate, and two SVD-compressed narrative components rank among the fifteen most influential features, confirming that narrative information contributes unique signal.

\textbf{Figures~\ref{fig:waterfall_high_risk} and \ref{fig:waterfall_low_risk}} illustrate local explanations for a high-risk and a low-risk patient, respectively. In both cases utilization history (prior inpatient visits and the heavy-utilizer score) provides the largest individual contributions.

\textbf{Figure~\ref{fig:shap_dependence}} visualizes SHAP dependence plots, revealing step-wise escalation of risk once prior inpatient visits reach two or more, and a transition from protective to risk-increasing contribution at roughly 12--13 concurrent medications.

\end{document}